\documentclass{article}

\usepackage[dblblindworkshop, final]{neurips_2025}

\workshoptitle{CogInterp: Interpreting Cognition in Deep Learning Models}

\usepackage[utf8]{inputenc} % allow utf-8 input
\usepackage[T1]{fontenc}    % use 8-bit T1 fonts
\usepackage{hyperref}       % hyperlinks
\usepackage{url}            % simple URL typesetting
\usepackage{booktabs}       % professional-quality tables
\usepackage{amsfonts}       % blackboard math symbols
\usepackage{nicefrac}       % compact symbols for 1/2, etc.
\usepackage{microtype}      % microtypography
\usepackage{xcolor}         % colors
\usepackage{graphicx}
\usepackage{amsmath}

\DeclareRobustCommand{\taustar}{\ensuremath{\overset{*}{\tau}}}

\title{Gradual Forgetting: Logarithmic Compression for Extending Transformer Context Windows}

\usepackage{authblk}

\author[1]{Billy Dickson}
\author[1,2]{Zoran Tiganj}

\affil[1]{Department of Computer Science, Indiana University Bloomington}
\affil[2]{Department of Psychological and Brain Sciences, Indiana University Bloomington}

\affil[ ]{\texttt{\{dicksonb, ztiganj\}@iu.edu}}

\begin{document}

\maketitle

\begin{abstract}
Most approaches to long-context processing increase the complexity of the transformer's internal architecture by integrating mechanisms such as recurrence or auxiliary memory modules. In this work, we introduce an alternative approach that modifies the input representation itself, rather than the transformer architecture. Inspired by cognitive models of human memory, our method applies a scale-invariant logarithmic compression to the input tokens. The resulting compressed representation is processed by a standard, unmodified transformer, preserving architectural simplicity. We evaluate this approach on the WikiText-103 and PG-19 language modeling benchmarks, showing a reduction in perplexity compared to uncompressed baselines. Moreover, performance improves consistently with longer compressed temporal contexts, showing that input‑level logarithmic compression is a simple and effective way to extend a transformer's long‑range memory.
\end{abstract}

\section{Introduction}

Transformers have become a dominant architecture for sequence modeling across domains such as language modeling, time-series forecasting, and dialog systems \citep{Vaswani+2017,lim2021temporal,devlin2018bert}, yet their ability to process long sequences is constrained by the quadratic complexity of self-attention \citep{child2019generating,beltagy2020longformer}. Existing solutions typically modify the model's architecture, employing segment-level recurrence \citep{dai2019transformer,rae2019compressive,NEURIPS2022_47e28862}, external memory modules \citep{graves2014neuralturingmachines,Weston2015,ko2024memreasoner,kang2025lm2largememorymodels}, or sparse and approximate attention mechanisms \citep{wu2022memorizing,child2019generating,beltagy2020longformer,zaheer2020bigbird,kitaev2020reformer,choromanski2021rethinking,wang2020linformer}, which often increases complexity and introduces state dependencies. In contrast, our work proposes a scale-invariant input transformation that compresses the input history before it reaches a standard transformer, requiring no architectural changes. Inspired by cognitive models of human memory that posit a logarithmic encoding of temporal information \citep{shankar2012scale,howard2014unified,tano2020local,findling2023brain,tiganj2019estimating,de1992gamma,grossberg1989neural} and their applications in deep neural networks \citep{jacques2021deepsith,jacques2022deep,mochizuki2024incorporating}, our method uses a bank of unimodal temporal filters to produce a log-compressed memory of the distant past. This compressed representation is concatenated with recent uncompressed tokens, preserving compatibility with existing attention mechanisms and supporting stateless batching. Evaluation on the WikiText-103 and PG-19 benchmarks shows that this approach improves perplexity over uncompressed baselines, with performance increasing as the compressed memory length grows.

\section{Model}
\begin{figure}[h!]
    \centering
    \begin{tabular}{lll}
    \textbf{A} &
    \textbf{B} &
    \textbf{C} \\
        {\includegraphics[width=0.31\textwidth]{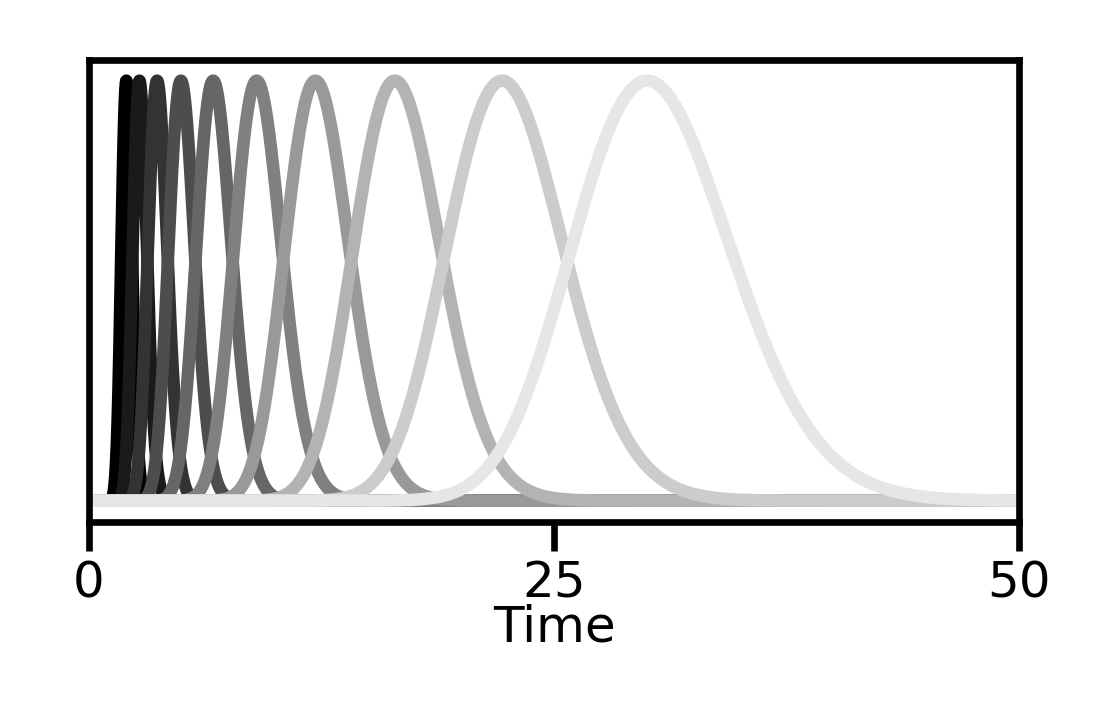}} &
        {\includegraphics[width=0.31\textwidth]{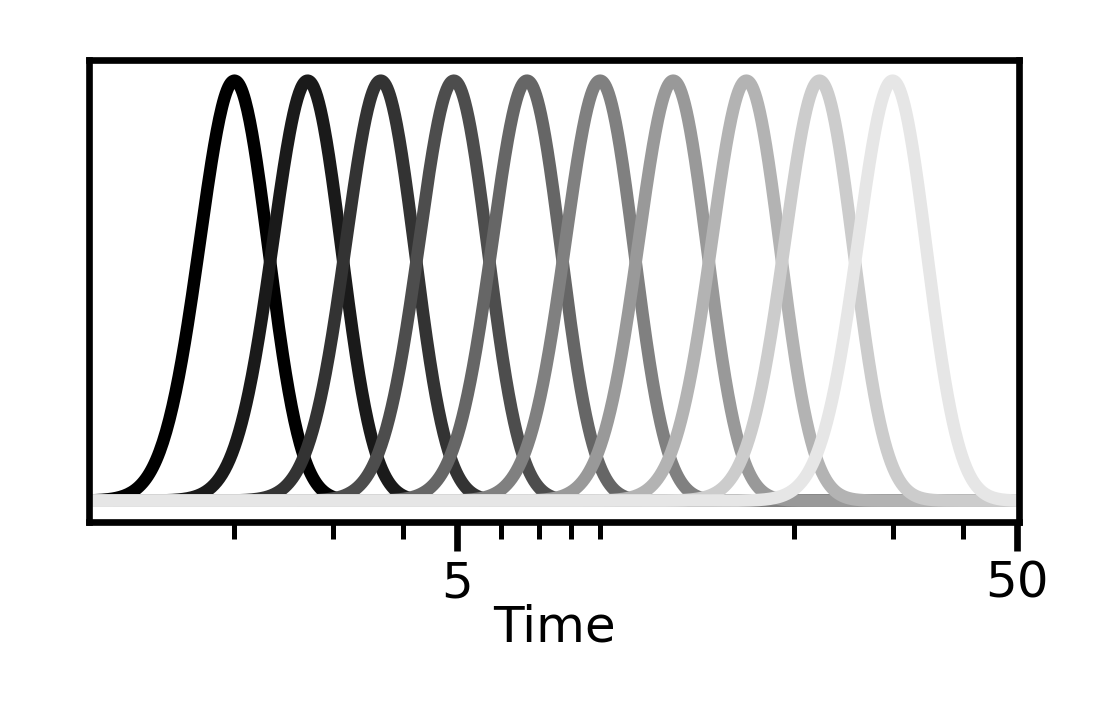}} &
        {\includegraphics[width=0.31\textwidth]{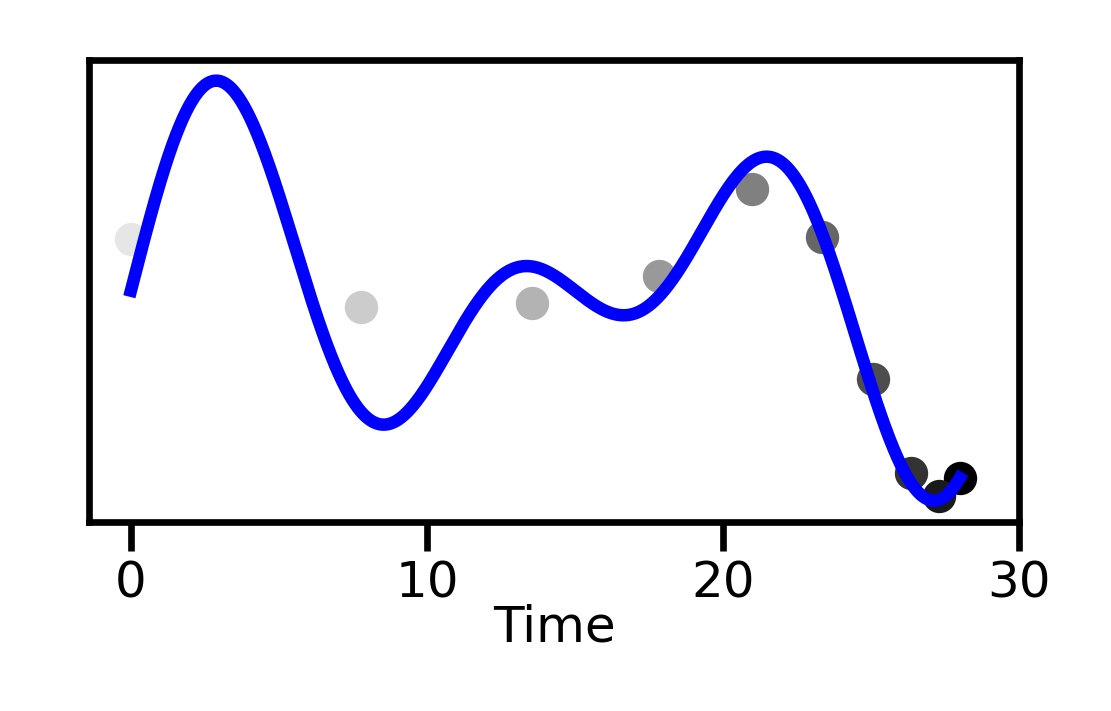}} 
    \end{tabular}
    \caption{Log-compressed impulse responses from ten $\tilde{f}$ neurons with $k=50$ show log-spaced peak times and a constant coefficient of variation (broader responses at later peaks). B. The same responses plotted on a log-time axis are uniformly spaced with equal widths. C. At time 30, $\tilde{f}$ neuron activations form a log-compressed memory of the input, approximating past values with finer resolution for more recent events. Grayscale dots correspond to impulse responses in panels A–B. Additional visualizations for $k=10$ and $k=100$ are shown in Fig.~\ref{fig:imp_res_appendix}.}

    \label{fig:imp_res_main}
\end{figure}

\begin{figure}[h!]
    \centering
    \begin{tabular}{ll}
    \textbf{A} &
    \textbf{\ \ \ \ \ \ \ \ \ \ \ B} \\
        {\includegraphics[width=0.6\textwidth]{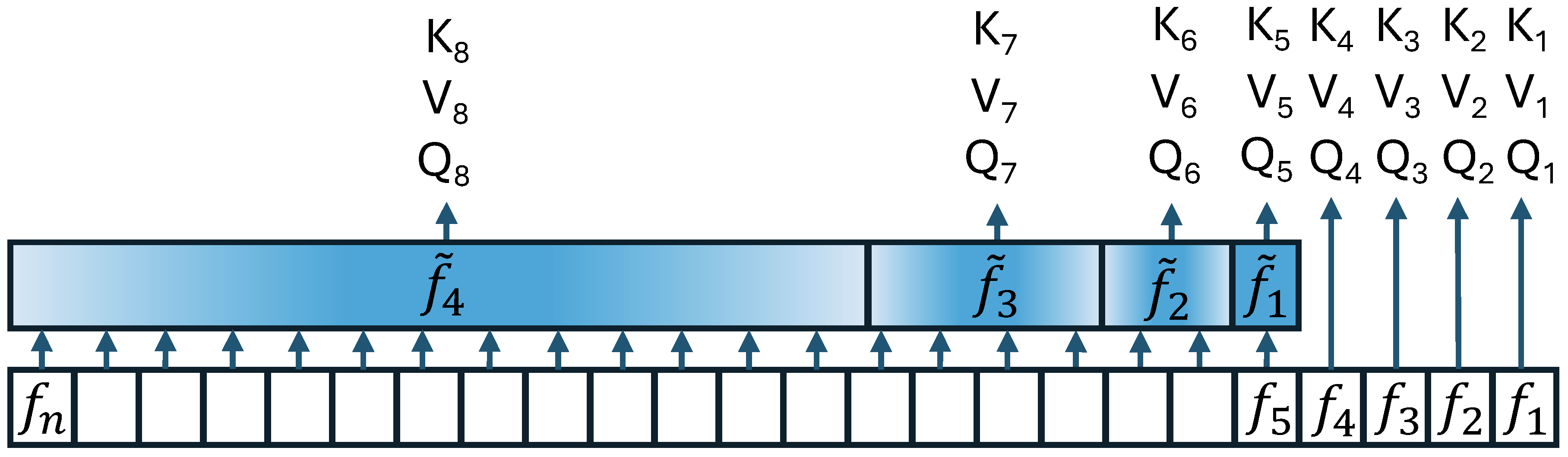}} & \ \ \ \ \ \ \ \ \ \ \ 
        {\includegraphics[width=0.212\textwidth]{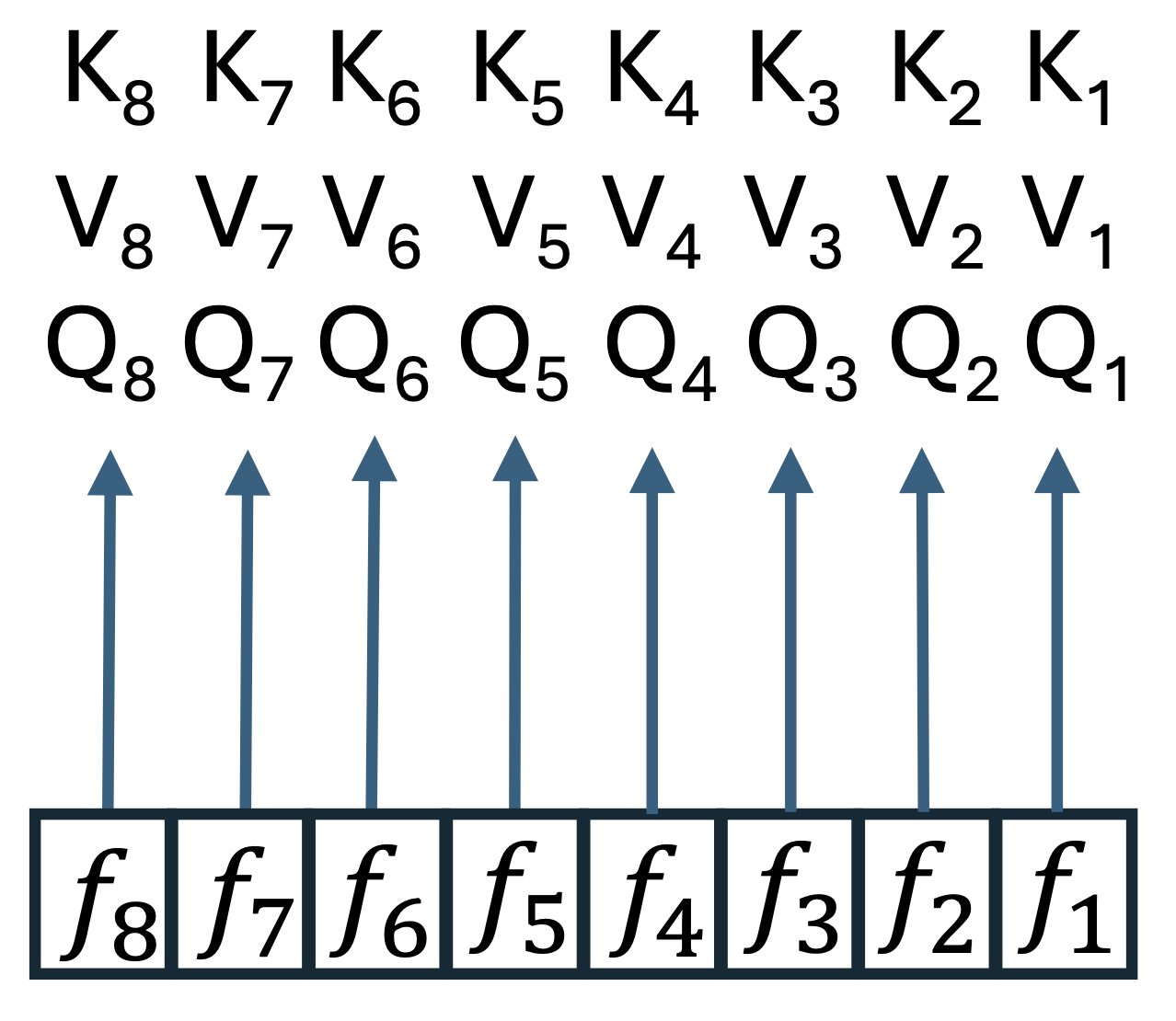}} 
    \end{tabular}
    \caption{\textbf{A.} Illustration of compressed memory applied to the input sequence. A subset of the last $m$ tokens, with $m=4$ in this example, is used to compute four keys, queries, and values. The rest of the input sequence is used as an input to the compressed memory composed of $L$ filters ($L=4$ in this example), producing an additional four keys, queries, and values. \textbf{B.} Illustration of a standard transformer where each token in the sequence is used to generate keys, queries, and values.  \label{fig:schematic_sequence}} 
\end{figure}

\begin{figure}[h!]
    \centering
    \includegraphics[width=0.75\textwidth]{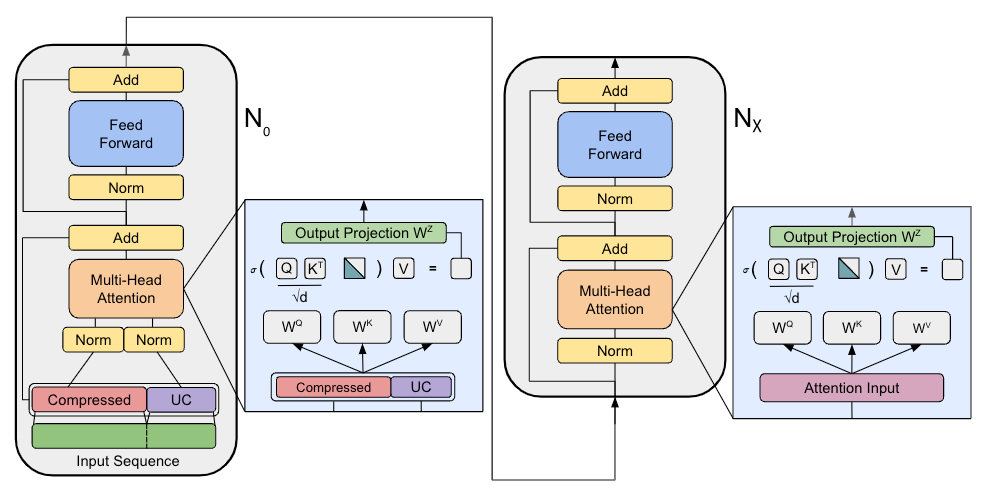}
    \caption{Architecture of the transformer based on scale-invariant compression. \emph{Compressed} represents the portion of the input that is compressed with the scale-invariant filters, \emph{UC} represents the uncompressed portion of the input. \emph{Norm} depicts Layer Normalization, \emph{Add} depicts residual connections, and \emph{Feed-Forward} depicts a multi-layer perceptron at the end of each layer. At the \emph{Multi-Head Attention} block in the first layer $N_0$, the output of the scale-invariant compression is concatenated with the uncompressed portion and projected to form queries, keys, and values. In subsequent layers $N_X$, where $X$ is the layer number, the input is passed through without additional compression.}
    \label{fig:first_block}
\end{figure}

We use a log‑spaced bank of scale‑invariant filters with impulse response $\Phi(t,\taustar)=\frac{k^{k+1}}{k!}\!\left(\frac{t}{\taustar}\right)^{\!k}\!e^{-k t/\taustar},$
where $k$ controls width and the peaks $\taustar_i$ are geometrically spaced (Fig.~\ref{fig:imp_res_main}), yielding rescaled copies that tile log‑time. Let $f(t)\in\mathbb{R}^d$ denote the token embedding at discrete step $t$. The compressed representation at time $t$ is obtained by a causal, depth‑wise 1‑D convolution:
\begin{equation}
    \tilde{f}(t,\taustar)=\sum_{t'=1}^{M}\Phi\!\big(t',\taustar\big)\,f(t-t').
    \label{eq:filtering}
\end{equation}
For each scale $\taustar$, $\tilde{f}(t,\taustar)$ is a smoothed, lagged estimate of $f(t-\taustar)$  (Fig.~\ref{fig:imp_res_main}C) termed Scale-Invariant Temporal History (SITH) \citep{jacques2021deepsith,jacques2022deep,shankar2012scale}. We truncate the impulse responses at a finite horizon $M$; in our experiments, we set $M=\taustar_{\max}$. All operations are vector-valued: the filter bank is applied independently to each embedding dimension (depth-wise 1-D convolution) to produce $L$  $d$‑dimensional ``compressed slots'' which are normalized via LayerNorm \citep{ba2016layernormalization} and concatenated with the $m$ most recent uncompressed tokens (Fig.~\ref{fig:schematic_sequence}) and subsequently processed by standard transformer layers (Fig.~\ref{fig:first_block}).

Forming the compressed memory costs $O(M L d)$ once per input chunk (depth‑wise 1‑D convolution over embeddings of size $d$).
Multi‑head attention then runs on a fixed length $m+L$ sequence with cost $O\big((m+L)^2 d\big)$ per layer, instead of $O\big((m+M)^2 d\big)$ if the entire history were attended directly.
Compression is applied only once as a preprocessing step before the first transformer block; subsequent layers operate on the concatenated sequence of fixed size. The training loss is computed only over the $m$ uncompressed tokens. This approach allows the model to efficiently leverage information from a much longer context without incurring the full quadratic cost of attention over the entire sequence.

\begin{figure}[!h]
    \centering
    \includegraphics[width=0.72\textwidth]{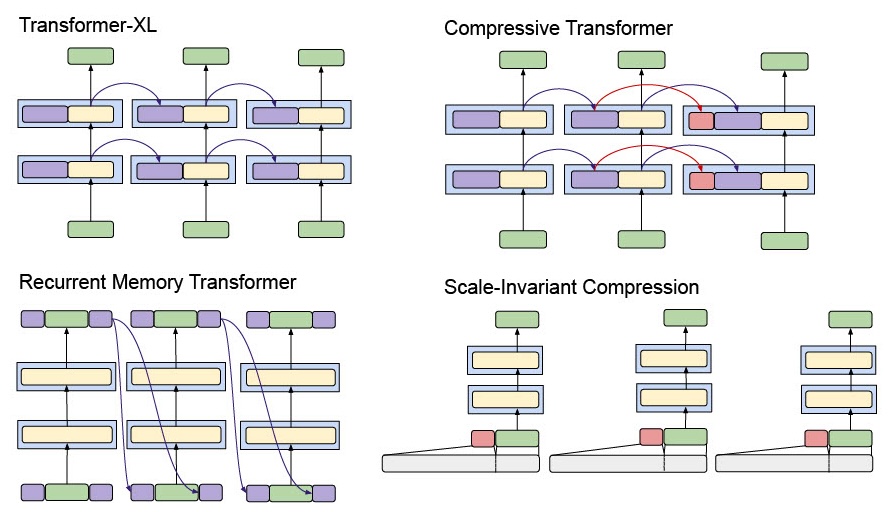}
    \caption{Comparison of long-context approaches. Transformer-XL, Compressive Transformer, and RMT handle long range by carrying state across segments (caching, compression, or recurrence). Our approach performs scale-invariant compression of the distant past to yield a fixed-size memory concatenated with recent tokens.}
    \label{fig:comparison}
\end{figure}

Fig.~\ref{fig:comparison} compares strategies for modeling long-range dependencies. Transformer-XL, Compressive Transformer, and Recurrent Memory Transformer process input sequentially in segments, passing state between them via caching, compression, or recurrence (purple and red memory blocks and arrows). Our scale-invariant compression instead preprocesses the input: a subset of the full input (gray) is logarithmically compressed (red) into a fixed-size representation and combined with recent uncompressed tokens (green). This joint input is then processed by a standard transformer (blue layers) without cross-segment state management.

\section{Training and Evaluation}
\label{sec:training}

Experiments were conducted on \textbf{WikiText-103}~\citep{merity2016pointer} and \textbf{PG-19}~\citep{rae2019compressive}. \textbf{WikiText-103} contains over 100M words from Wikipedia, while \textbf{PG-19} comprises 28k Project Gutenberg books (1.9B words) exceeding 100k words each, designed for long-context modeling.

We trained transformer models and examined the effects of compression parameter $k$, and number of filters $L$, using an uncompressed sequence length of 256 tokens and $k \in \{100,150,200\}$. We fix the geometric spacing constant to $c=0.19$. With $\taustar_{\min}=1$, the filter peaks are
\[
\taustar_i=\taustar_{\min}(1+c)^{i-1},\quad i=1,\dots,L,
\]
which implies $\taustar_{\max}=\taustar_{\min}(1+c)^{L-1}$. Thus increasing $L$ exponentially increases the temporal span (e.g., $L=53\Rightarrow \taustar_{\max}\approx 8192$ tokens). In our sweeps we varied $L\in\{5,9,\ldots,53\}$, corresponding to filter windows from 2 to 8192 tokens.

For each $k$, we trained 13 models with $L$ log-spaced filters (5–53), spanning window sizes from 2–8192 tokens. Each setting included a compression model (Fig.~\ref{fig:schematic_sequence}A) and a delta-pulse control (Fig.~\ref{fig:schematic_sequence}B) that bypassed compression via shift-register buffers, equivalent to retaining embeddings of the $L$ preceding tokens. All models followed the GPT-2 Small architecture~\citep{Radford2019LanguageMA} (12 layers, 12 heads, 768-d embeddings, 3{,}072-d MLP, 50{,}304 vocab (50{,}257 rounded up to the nearest multiple of 64 for training efficiency); $\sim$124M parameters) with minor variation in number of parameters due to size of the learned positional encoding. Training used AdamW ($\beta$={0.9,0.95}), weight decay 0.1, linear warmup for 700 steps to a peak learning rate of $6\times10^{-4}$ with cosine decay to $6\times10^{-5}$, batch size $\sim$16{,}384 tokens/step, gradient clipping 1.0, and no dropout. Each model trained $\sim$48h on a 40 GB A100 GPU (90 epochs on WikiText-103; 4 on PG-19). We used a sliding window with stride $m$ when forming batches, computing loss only over the $m$ uncompressed tokens at each step. We report \emph{raw perplexity} as the perplexity computed using the GPT-2 BPE tokenizer, given by \( e^{\text{Loss}_{\mu}} \), where \emph{\( \text{Loss}_{\mu} \)} denotes the average cross-entropy over the tokenized train or test subset. Following \citet{rae2019compressive} we report \emph{per-word perplexity} as \( e^{\text{Loss}_{tot} / n_{\text{words}}} \), where \emph{\( \text{Loss}_{\text{tot}} \)} denotes the total cross-entropy over the tokenized subset and \( n_{\text{words}} \) is the total number of words in the given subset. This facilitates direct comparisons between our GPT-2 BPE tokenizer-based models and other models using whitespace tokenization.

\section{Results}

The proposed scale-invariant compression filters consistently improve model performance by efficiently extending the temporal context on both the WikiText-103 and PG-19 datasets. This approach outperforms a control model using simple delta pulse filters, particularly as the number of memory filters $L$ increases. As shown in Fig.~\ref{fig:wikitext_results}, test perplexity on both datasets generally decreases as the number of filters 
$L$ increases, with small non‑monotonicities. Because the filter peaks are geometrically spaced, this corresponds to an approximately log‑scale dependence on the maximum filter peak time (see Appendix~\ref{sec:appendix_results} for numerical values of perplexity in each tested condition). This improvement becomes most apparent when the temporal context window significantly exceeds the number of filters (e.g., for $L>17$ on WikiText-103 and $L>21$ on PG-19), underscoring the benefit of capturing long-range dependencies. 

\begin{figure*}[!h]
    \centering
    \begin{tabular}{ll}
    \textbf{A} &
    \textbf{B} \\
        {\includegraphics[width=0.47\textwidth]{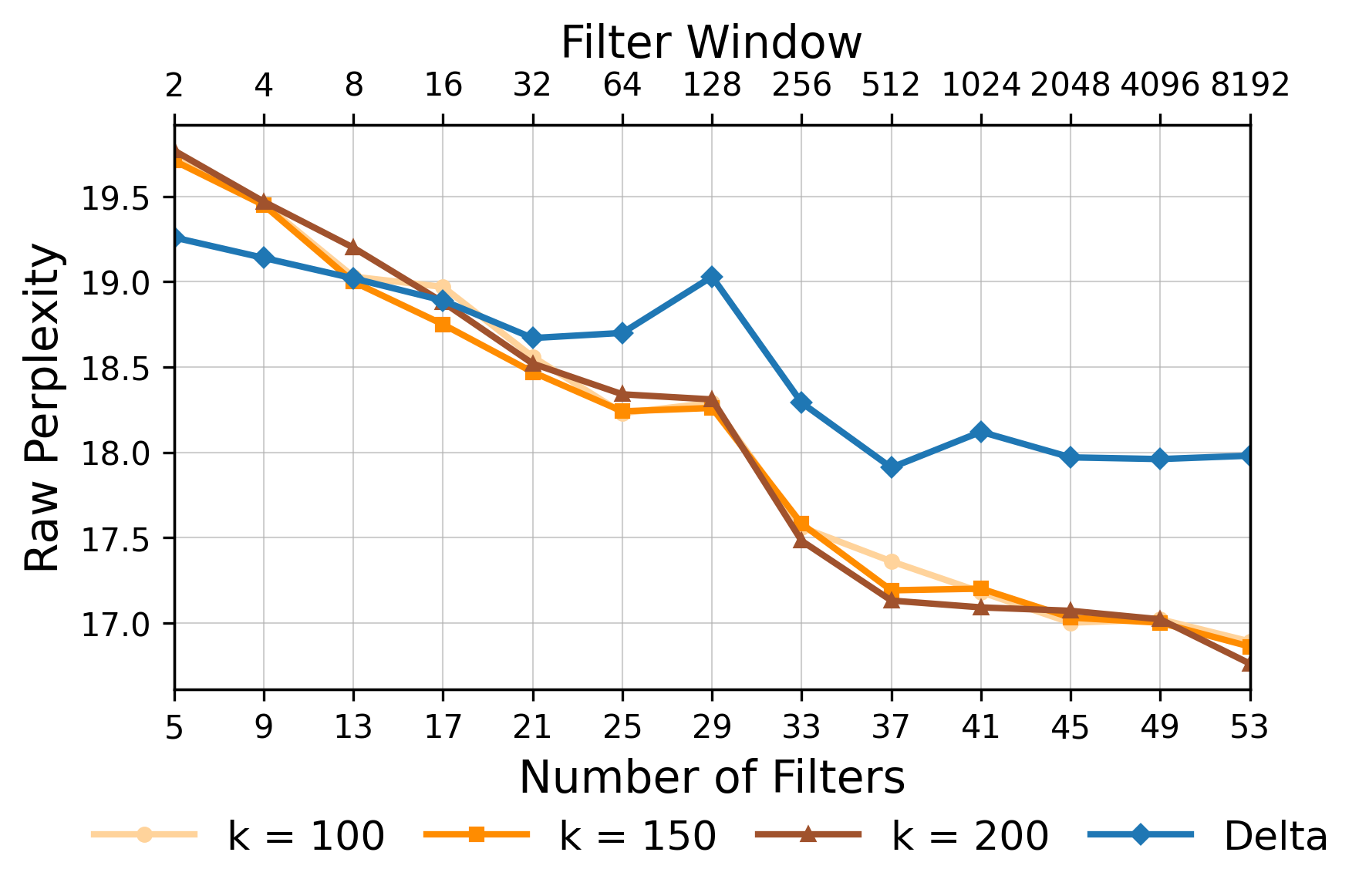}} &
        {\includegraphics[width=0.47\textwidth]{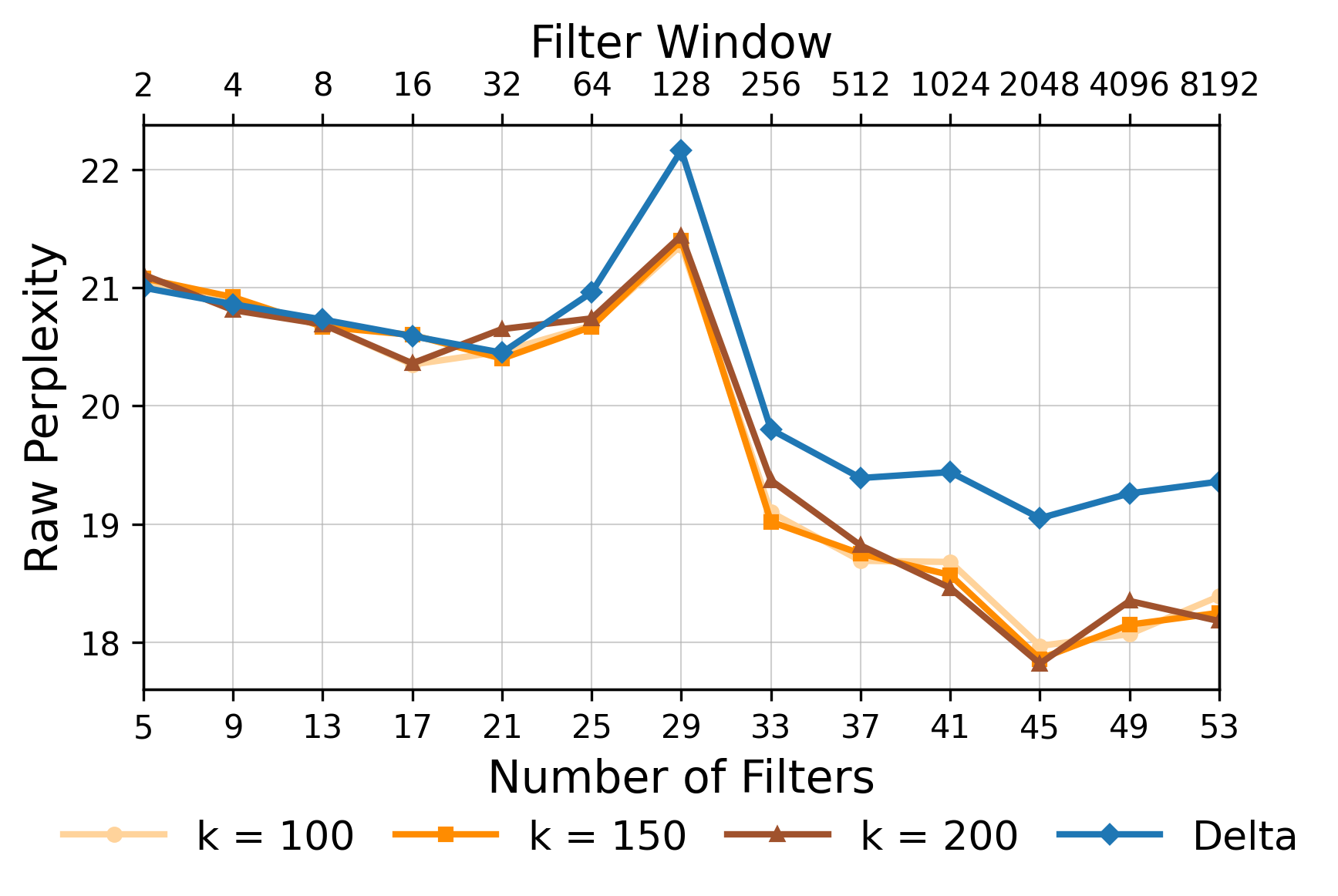}} 
    \end{tabular}
    \caption{Test set perplexity decreases as the number of filters $L$ increases. WikiText-103 (A) shows an approximately linear decline; PG-19 (B) shows a downward, step-like trend with a knee around mid-range $L$. The top axis indicates the corresponding filter window $\tau_{\max}$ (geometric spacing). The results are consistent for different values of $k$. }
    \label{fig:wikitext_results}
\end{figure*}

On WikiText-103, our best model achieved a per-word perplexity of \textbf{23.56} (with $L=53$ filters covering an 8192-token window). As detailed in Table~\ref{tab:model_comparison}, this result is competitive with other transformer architectures utilizing long-context memory approaches with similar parameter counts. While these results are promising, we acknowledge that models with significantly more parameters (over 200M) achieve lower perplexity scores. The performance trend was replicated on the PG-19 dataset, though a direct comparison with other models was not possible due to a lack of published results at a similar scale or with per-word normalization of perplexity.

\begin{table}[!h]
\centering
\begin{tabular}{lcccc}
    \toprule
    \textbf{Model} & \textbf{\# param} & \textbf{Attn. Length} & \textbf{Per-word PPL} \\
    \midrule
    Delta Pulse Filters (Our control model)                                     & 124M & 309 & 25.48 \\
    Transformer-XL Standard \citep{dai2019transformer}                          & 151M & 300 & 24.00 \\
    RMT \citep{NEURIPS2022_47e28862}                                            & 151M & 175 & 24.85 \\
    Transformer-XL + RMT \citep{NEURIPS2022_47e28862}                           & 151M & 310 & 23.99 \\
    Scale-Invariant Compression Filters (Ours)                                  & 124M & 309 & \textbf{23.56} \\
    \bottomrule
\end{tabular}
\vspace{4pt}
\caption{Comparison of transformer models of similar size (\emph{\# param}) on the WikiText-103 test set. Our model achieves the lowest per-word perplexity. \emph{Attn. Length} is the input size considered by attention ($m+L$ for our model, with $m=256$ uncompressed tokens plus $L=53$ compressed inputs giving the total of 309 tokens).}
\label{tab:model_comparison}
\end{table}

\section{Discussion}

Unlike transformers, the human brain does not store a verbatim record of linguistic input. Instead, it learns statistical regularities in language by dynamically maintaining a memory representation of the recent past~\citep{saffran1996statistical,kuhl2004early}. Cognitive scientists have argued that this representation forms a \emph{mental timeline} of the past~\citep{brown2007temporal,howard2015distributed}. Consistent with scale-invariant power-law decay of memory~\citep{ebbinghaus2013memory,wixted2004psychology} and the Weber--Fechner law~\citep{fechner1860elemente}, this timeline is thought to have a temporal resolution that gradually decreases from recent to distant events, yielding a logarithmically compressed representation. Neuroscience studies support this view, reporting neurons that activate sequentially with logarithmically compressed temporal receptive fields~\citep{cao2022internally,tiganj2018compressed,eichenbaum2014time}. Our approach augments transformer input representations with such a memory timeline, enabling the model to capture temporal dependencies across a wide input range. Logarithmic compression implies that the effective temporal range expands exponentially with the number of attention scores, offering a resource-efficient representation. Moreover, it entails that the density of neurons decreases as a power-law function of the peak times of their receptive fields. Power-law decay of long-range correlations is a pervasive phenomenon observed in DNA sequences~\citep{mantegna1994linguistic}, musical rhythm spectra~\citep{levitin2012musical}, earthquake statistics~\citep{abe2004scale}, and natural language~\citep{ebeling1995long,altmann2012origin}. \citet{lin2017critical} similarly showed that mutual information between symbols can decay as a power law with increasing separation in context-free grammars. Integrating transformer architectures with power-law--decaying memory thus provides a principled framework for modeling domains where such statistical regularities naturally arise. A complementary, neurally motivated \emph{time-local transformer} builds a log-compressed timeline via fixed recurrent dynamics and attends to it at each step, trading higher per-step compute for strict time-locality and greater biological plausibility \citep{dickson2025time}.

\section{Conclusion}
Our approach augments transformers with a scale-invariant memory inspired by a cognitive model of human memory. By incorporating temporal logarithmic compression as an input preprocessing step, our model efficiently encodes long-range dependencies into a fixed-size representation, allowing a standard transformer architecture to capture these dependencies effectively while maintaining computational tractability. Through experiments on the WikiText-103 and PG-19 datasets, we demonstrated that our model outperforms other transformer models of similar size. The observed gradual decrease in perplexity with increased temporal context highlights the efficacy of the scale-invariant memory representation in capturing long-range correlations. These findings illustrate that integrating cognitive principles into neural architectures can lead to more efficient language models.  

\newpage
\bibliography{references}

\begin{thebibliography}{47}
\providecommand{\natexlab}[1]{#1}
\providecommand{\url}[1]{\texttt{#1}}
\expandafter\ifx\csname urlstyle\endcsname\relax
  \providecommand{\doi}[1]{doi: #1}\else
  \providecommand{\doi}{doi: \begingroup \urlstyle{rm}\Url}\fi

\bibitem[Abe \& Suzuki(2004)Abe and Suzuki]{abe2004scale}
Sumiyoshi Abe and Norikazu Suzuki.
\newblock Scale-free network of earthquakes.
\newblock \emph{Europhysics Letters}, 65\penalty0 (4):\penalty0 581--586, 2004.

\bibitem[Altmann et~al.(2012)Altmann, Cristadoro, and Esposti]{altmann2012origin}
Eduardo~G Altmann, Giampaolo Cristadoro, and Mirko~Degli Esposti.
\newblock On the origin of long-range correlations in texts.
\newblock \emph{Proceedings of the National Academy of Sciences}, 109\penalty0 (29):\penalty0 11582--11587, 2012.

\bibitem[Ba et~al.(2016)Ba, Kiros, and Hinton]{ba2016layernormalization}
Jimmy~Lei Ba, Jamie~Ryan Kiros, and Geoffrey~E Hinton.
\newblock Layer normalization.
\newblock \emph{arXiv preprint arXiv:1607.06450}, 2016.

\bibitem[Beltagy et~al.(2020)Beltagy, Peters, and Cohan]{beltagy2020longformer}
Iz~Beltagy, Matthew~E Peters, and Arman Cohan.
\newblock Longformer: The long-document transformer.
\newblock \emph{arXiv preprint arXiv:2004.05150}, 2020.

\bibitem[Brown et~al.(2007)Brown, Neath, and Chater]{brown2007temporal}
Gordon~DA Brown, Ian Neath, and Nick Chater.
\newblock A temporal ratio model of memory.
\newblock \emph{Psychological review}, 114\penalty0 (3):\penalty0 539, 2007.

\bibitem[Bulatov et~al.(2022)Bulatov, Kuratov, and Burtsev]{NEURIPS2022_47e28862}
Aydar Bulatov, Yury Kuratov, and Mikhail Burtsev.
\newblock Recurrent memory transformer.
\newblock \emph{Advances in Neural Information Processing Systems}, 35:\penalty0 11079--11091, 2022.

\bibitem[Cao et~al.(2022)Cao, Bladon, Charczynski, Hasselmo, and Howard]{cao2022internally}
Rui Cao, John~H Bladon, Stephen~J Charczynski, Michael~E Hasselmo, and Marc~W Howard.
\newblock Internally generated time in the rodent hippocampus is logarithmically compressed.
\newblock \emph{Elife}, 11:\penalty0 e75353, 2022.

\bibitem[Child et~al.(2019)Child, Gray, Radford, and Sutskever]{child2019generating}
Rewon Child, Scott Gray, Alec Radford, and Ilya Sutskever.
\newblock Generating long sequences with sparse transformers.
\newblock \emph{Advances in Neural Information Processing Systems}, pp.\  11969--11979, 2019.

\bibitem[Choromanski et~al.(2021)Choromanski, Likhosherstov, Dohan, Song, Gane, Sarlos, Hawkins, Davis, Mohiuddin, Kaiser, et~al.]{choromanski2021rethinking}
Krzysztof Choromanski, Valerii Likhosherstov, David Dohan, Xingyou Song, Andreea Gane, Tamas Sarlos, Peter Hawkins, Jared Davis, Afroz Mohiuddin, Lukasz Kaiser, et~al.
\newblock Rethinking attention with performers.
\newblock In \emph{International Conference on Learning Representations}, 2021.

\bibitem[Dai et~al.(2019)Dai, Yang, Yang, Carbonell, Le, and Salakhutdinov]{dai2019transformer}
Zihang Dai, Zhilin Yang, Yiming Yang, Jaime Carbonell, Quoc~V Le, and Ruslan Salakhutdinov.
\newblock Transformer-xl: Attentive language models beyond a fixed-length context.
\newblock In \emph{Proceedings of the 57th Annual Meeting of the Association for Computational Linguistics}, pp.\  2978--2988, 2019.

\bibitem[De~Vries \& Principe(1992)De~Vries and Principe]{de1992gamma}
Bert De~Vries and Jose~C Principe.
\newblock The gamma model—a new neural model for temporal processing.
\newblock \emph{Neural networks}, 5\penalty0 (4):\penalty0 565--576, 1992.

\bibitem[Devlin et~al.(2019)Devlin, Chang, Lee, and Toutanova]{devlin2018bert}
Jacob Devlin, Ming-Wei Chang, Kenton Lee, and Kristina Toutanova.
\newblock Bert: Pre-training of deep bidirectional transformers for language understanding.
\newblock In \emph{Proceedings of the 2019 conference of the North American chapter of the association for computational linguistics: human language technologies, volume 1}, pp.\  4171--4186, 2019.

\bibitem[Dickson et~al.(2025)Dickson, Mochizuki-Freeman, Kabir, and Tiganj]{dickson2025time}
Billy Dickson, James Mochizuki-Freeman, Md~Rysul Kabir, and Zoran Tiganj.
\newblock Time-local transformer.
\newblock \emph{Computational Brain \& Behavior}, pp.\  1--13, 2025.

\bibitem[Ebbinghaus(2013)]{ebbinghaus2013memory}
Hermann Ebbinghaus.
\newblock Memory: A contribution to experimental psychology.
\newblock \emph{Annals of neurosciences}, 20\penalty0 (4):\penalty0 155, 2013.

\bibitem[Ebeling \& Neiman(1995)Ebeling and Neiman]{ebeling1995long}
Werner Ebeling and Alexander Neiman.
\newblock Long-range correlations between letters and sentences in texts.
\newblock \emph{Physica A: Statistical Mechanics and its Applications}, 215\penalty0 (3):\penalty0 233--241, 1995.

\bibitem[Eichenbaum(2014)]{eichenbaum2014time}
Howard Eichenbaum.
\newblock Time cells in the hippocampus: a new dimension for mapping memories.
\newblock \emph{Nature Reviews Neuroscience}, 15\penalty0 (11):\penalty0 732--744, 2014.

\bibitem[Fechner(1860)]{fechner1860elemente}
Gustav~Theodor Fechner.
\newblock \emph{Elemente der psychophysik}, volume~2.
\newblock Breitkopf u. H{\"a}rtel, 1860.

\bibitem[Findling et~al.(2025)Findling, Hubert, Laboratory, Acerbi, Benson, Benson, Birman, Bonacchi, Buchanan, Bruijns, et~al.]{findling2023brain}
Charles Findling, Felix Hubert, International~Brain Laboratory, Luigi Acerbi, Brandon Benson, Julius Benson, Daniel Birman, Niccol{\`o} Bonacchi, E~Kelly Buchanan, Sebastian Bruijns, et~al.
\newblock Brain-wide representations of prior information in mouse decision-making.
\newblock \emph{Nature}, 645\penalty0 (8079):\penalty0 192--200, 2025.

\bibitem[Graves et~al.(2014)Graves, Wayne, and Danihelka]{graves2014neuralturingmachines}
Alex Graves, Greg Wayne, and Ivo Danihelka.
\newblock Neural turing machines.
\newblock \emph{arXiv preprint arXiv:1410.5401}, 2014.

\bibitem[Grossberg \& Schmajuk(1989)Grossberg and Schmajuk]{grossberg1989neural}
Stephen Grossberg and Nestor~A Schmajuk.
\newblock Neural dynamics of adaptive timing and temporal discrimination during associative learning.
\newblock \emph{Neural networks}, 2\penalty0 (2):\penalty0 79--102, 1989.

\bibitem[Howard et~al.(2014)Howard, MacDonald, Tiganj, Shankar, Du, Hasselmo, and Eichenbaum]{howard2014unified}
Marc~W Howard, Christopher~J MacDonald, Zoran Tiganj, Karthik~H Shankar, Qian Du, Michael~E Hasselmo, and Howard Eichenbaum.
\newblock A unified mathematical framework for coding time, space, and sequences in the hippocampal region.
\newblock \emph{Journal of Neuroscience}, 34\penalty0 (13):\penalty0 4692--4707, 2014.

\bibitem[Howard et~al.(2015)Howard, Shankar, Aue, and Criss]{howard2015distributed}
Marc~W. Howard, Karthik~H. Shankar, William~R. Aue, and Amy~H. Criss.
\newblock A distributed representation of internal time.
\newblock \emph{Psychological Review}, 122\penalty0 (1):\penalty0 24--53, 2015.

\bibitem[Jacques et~al.(2021)Jacques, Tiganj, Howard, and Sederberg]{jacques2021deepsith}
Brandon Jacques, Zoran Tiganj, Marc Howard, and Per~B Sederberg.
\newblock Deepsith: Efficient learning via decomposition of what and when across time scales.
\newblock \emph{Advances in Neural Information Processing Systems}, 34:\penalty0 27530--27541, 2021.

\bibitem[Jacques et~al.(2022)Jacques, Tiganj, Sarkar, Howard, and Sederberg]{jacques2022deep}
Brandon~G Jacques, Zoran Tiganj, Aakash Sarkar, Marc Howard, and Per Sederberg.
\newblock A deep convolutional neural network that is invariant to time rescaling.
\newblock In \emph{International conference on machine learning}, pp.\  9729--9738. PMLR, 2022.

\bibitem[Kang et~al.(2025)Kang, Wu, Christianos, Chan, Greenlee, Thomas, Purtorab, and Toulis]{kang2025lm2largememorymodels}
Jikun Kang, Wenqi Wu, Filippos Christianos, Alex~J Chan, Fraser Greenlee, George Thomas, Marvin Purtorab, and Andy Toulis.
\newblock Lm2: Large memory models.
\newblock \emph{arXiv preprint arXiv:2502.06049}, 2025.

\bibitem[Kitaev et~al.(2020)Kitaev, Kaiser, and Levskaya]{kitaev2020reformer}
Nikita Kitaev, {\L}ukasz Kaiser, and Anselm Levskaya.
\newblock Reformer: The efficient transformer.
\newblock In \emph{International Conference on Learning Representations}, 2020.

\bibitem[Ko et~al.(2024)Ko, Dai, Das, Kollias, Chaudhury, and Lozano]{ko2024memreasoner}
Ching-Yun Ko, Sihui Dai, Payel Das, Georgios Kollias, Subhajit Chaudhury, and Aurelie Lozano.
\newblock Memreasoner: A memory-augmented {LLM} architecture for multi-hop reasoning.
\newblock In \emph{The First Workshop on System-2 Reasoning at Scale, NeurIPS'24}, 2024.

\bibitem[Kuhl(2004)]{kuhl2004early}
Patricia~K Kuhl.
\newblock Early language acquisition: cracking the speech code.
\newblock \emph{Nature reviews neuroscience}, 5\penalty0 (11):\penalty0 831--843, 2004.

\bibitem[Levitin et~al.(2012)Levitin, Chordia, and Menon]{levitin2012musical}
Daniel~J Levitin, Parag Chordia, and Vinod Menon.
\newblock Musical rhythm spectra from bach to joplin obey a 1/f power law.
\newblock \emph{Proceedings of the National Academy of Sciences}, 109\penalty0 (10):\penalty0 3716--3720, 2012.

\bibitem[Lim et~al.(2021)Lim, Arik, Loeff, and Pfister]{lim2021temporal}
Bryan Lim, Sercan~O Arik, Nicolas Loeff, and Tomas Pfister.
\newblock Temporal fusion transformers for interpretable multi-horizon time series forecasting.
\newblock \emph{International Journal of Forecasting}, 2021.

\bibitem[Lin \& Tegmark(2017)Lin and Tegmark]{lin2017critical}
Henry~W Lin and Max Tegmark.
\newblock Critical behavior in physics and probabilistic formal languages.
\newblock \emph{Entropy}, 19\penalty0 (7):\penalty0 299, 2017.

\bibitem[Mantegna et~al.(1994)Mantegna, Buldyrev, Goldberger, Havlin, Peng, Simons, and Stanley]{mantegna1994linguistic}
Rosario~N Mantegna, Sergey~V Buldyrev, Ary~L Goldberger, Shlomo Havlin, Chung-Kang Peng, M~Simons, and H~Eugene Stanley.
\newblock Linguistic features of noncoding dna sequences.
\newblock \emph{Physical review letters}, 73\penalty0 (23):\penalty0 3169--3172, 1994.

\bibitem[Merity et~al.(2016)Merity, Xiong, Bradbury, and Socher]{merity2016pointer}
Stephen Merity, Caiming Xiong, James Bradbury, and Richard Socher.
\newblock Pointer sentinel mixture models.
\newblock \emph{arXiv preprint arXiv:1609.07843}, 2016.

\bibitem[Mochizuki-Freeman et~al.(2024)Mochizuki-Freeman, Kabir, and Tiganj]{mochizuki2024incorporating}
James Mochizuki-Freeman, Md~Rysul Kabir, and Zoran Tiganj.
\newblock Incorporating a cognitive model for evidence accumulation into deep reinforcement learning agents.
\newblock In \emph{Proceedings of the Annual Meeting of the Cognitive Science Society}, volume~46, 2024.

\bibitem[Radford et~al.(2019)Radford, Wu, Child, Luan, Amodei, Sutskever, et~al.]{Radford2019LanguageMA}
Alec Radford, Jeffrey Wu, Rewon Child, David Luan, Dario Amodei, Ilya Sutskever, et~al.
\newblock Language models are unsupervised multitask learners.
\newblock \emph{OpenAI blog}, 1\penalty0 (8):\penalty0 9, 2019.

\bibitem[Rae et~al.(2019)Rae, Potapenko, Jayakumar, and Lillicrap]{rae2019compressive}
Jack~W Rae, Anna Potapenko, Siddhant~M Jayakumar, and Timothy~P Lillicrap.
\newblock Compressive transformers for long-range sequence modelling.
\newblock \emph{arXiv preprint arXiv:1911.05507}, 2019.

\bibitem[Saffran et~al.(1996)Saffran, Aslin, and Newport]{saffran1996statistical}
Jenny~R Saffran, Richard~N Aslin, and Elissa~L Newport.
\newblock Statistical learning by 8-month-old infants.
\newblock \emph{Science}, 274\penalty0 (5294):\penalty0 1926--1928, 1996.

\bibitem[Shankar \& Howard(2012)Shankar and Howard]{shankar2012scale}
Karthik~H Shankar and Marc~W Howard.
\newblock A scale-invariant internal representation of time.
\newblock \emph{Neural Computation}, 24\penalty0 (1):\penalty0 134--193, 2012.

\bibitem[Tano et~al.(2020)Tano, Dayan, and Pouget]{tano2020local}
Pablo Tano, Peter Dayan, and Alexandre Pouget.
\newblock A local temporal difference code for distributional reinforcement learning.
\newblock \emph{Advances in neural information processing systems}, 33:\penalty0 13662--13673, 2020.

\bibitem[Tiganj et~al.(2018)Tiganj, Cromer, Roy, Miller, and Howard]{tiganj2018compressed}
Zoran Tiganj, Jason~A Cromer, Jefferson~E Roy, Earl~K Miller, and Marc~W Howard.
\newblock Compressed timeline of recent experience in monkey lateral prefrontal cortex.
\newblock \emph{Journal of cognitive neuroscience}, 30\penalty0 (7):\penalty0 935--950, 2018.

\bibitem[Tiganj et~al.(2019)Tiganj, Gershman, Sederberg, and Howard]{tiganj2019estimating}
Zoran Tiganj, Samuel~J Gershman, Per~B Sederberg, and Marc~W Howard.
\newblock Estimating scale-invariant future in continuous time.
\newblock \emph{Neural Computation}, 31\penalty0 (4):\penalty0 681--709, 2019.

\bibitem[Vaswani et~al.(2017)Vaswani, Shazeer, Parmar, Uszkoreit, Jones, Gomez, Kaiser, and Polosukhin]{Vaswani+2017}
Ashish Vaswani, Noam Shazeer, Niki Parmar, Jakob Uszkoreit, Llion Jones, Aidan~N Gomez, {\L}~ukasz Kaiser, and Illia Polosukhin.
\newblock Attention is all you need.
\newblock \emph{Advances in Neural Information Processing Systems}, 30, 2017.

\bibitem[Wang et~al.(2020)Wang, Li, Khabsa, Fang, and Ma]{wang2020linformer}
Sinong Wang, Belinda~Z Li, Madian Khabsa, Han Fang, and Hao Ma.
\newblock Linformer: Self-attention with linear complexity.
\newblock In \emph{Advances in Neural Information Processing Systems}, 2020.

\bibitem[Weston et~al.(2015)Weston, Chopra, and Bordes]{Weston2015}
Jason Weston, Sumit Chopra, and Antoine Bordes.
\newblock Memory networks.
\newblock \emph{3rd International Conference on Learning Representations, ICLR 2015 - Conference Track Proceedings}, 2015.

\bibitem[Wixted(2004)]{wixted2004psychology}
John~T Wixted.
\newblock The psychology and neuroscience of forgetting.
\newblock \emph{Annu. Rev. Psychol.}, 55\penalty0 (1):\penalty0 235--269, 2004.

\bibitem[Wu et~al.(2022)Wu, Rabe, Hutchins, and Szegedy]{wu2022memorizing}
Yuhuai Wu, Markus~Norman Rabe, DeLesley Hutchins, and Christian Szegedy.
\newblock Memorizing transformers.
\newblock In \emph{International Conference on Learning Representations}, 2022.

\bibitem[Zaheer et~al.(2020)Zaheer, Guruganesh, Dubey, Ainslie, Alberti, Ontanon, Pham, Ravula, Wang, Yang, et~al.]{zaheer2020bigbird}
Manzil Zaheer, Guru Guruganesh, Kumar~Avinava Dubey, Joshua Ainslie, Chris Alberti, Santiago Ontanon, Philip Pham, Anirudh Ravula, Qifan Wang, Li~Yang, et~al.
\newblock Big bird: Transformers for longer sequences.
\newblock \emph{Advances in Neural Information Processing Systems}, 33:\penalty0 17283--17297, 2020.

\end{thebibliography}
\bibliographystyle{references}

\renewcommand{\thetable}{A\arabic{table}}
\renewcommand{\thefigure}{A\arabic{figure}}
\setcounter{table}{0}
\setcounter{figure}{0}

\clearpage

\appendix

\section{Additional visualizations and results}
\label{sec:appendix_results}

\begin{figure}[h!]
    \centering
    \begin{tabular}{lll}
    \textbf{D} &
    \textbf{E} &
    \textbf{F} \\
        {\includegraphics[width=0.31\textwidth]{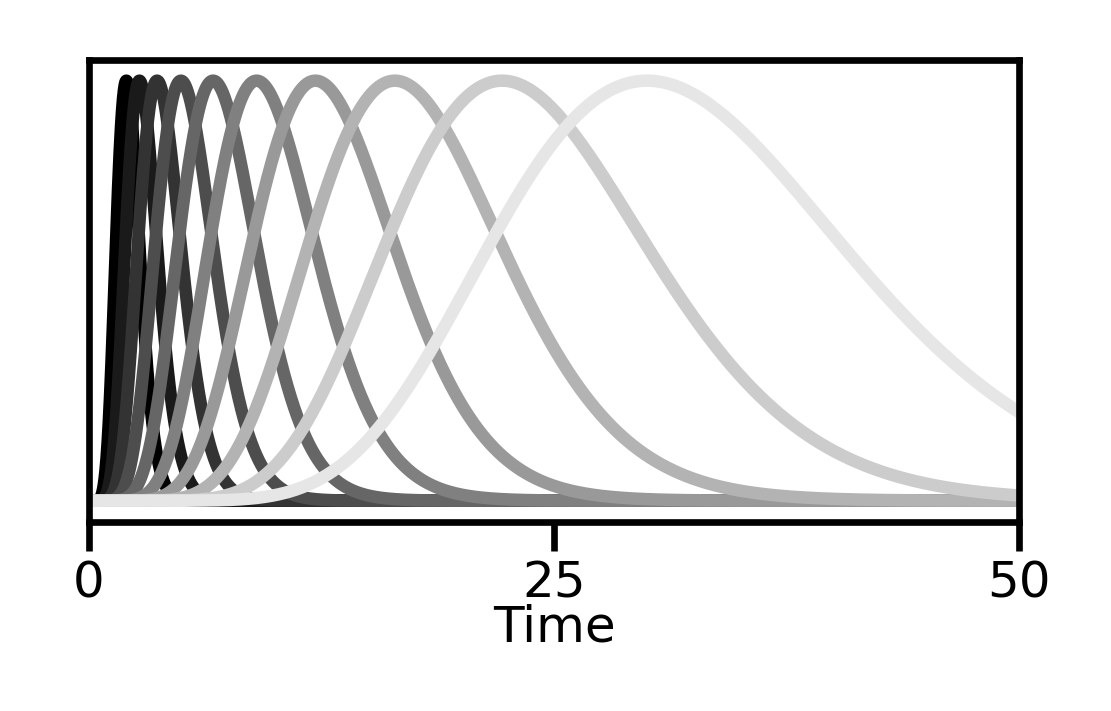}} &
        {\includegraphics[width=0.31\textwidth]{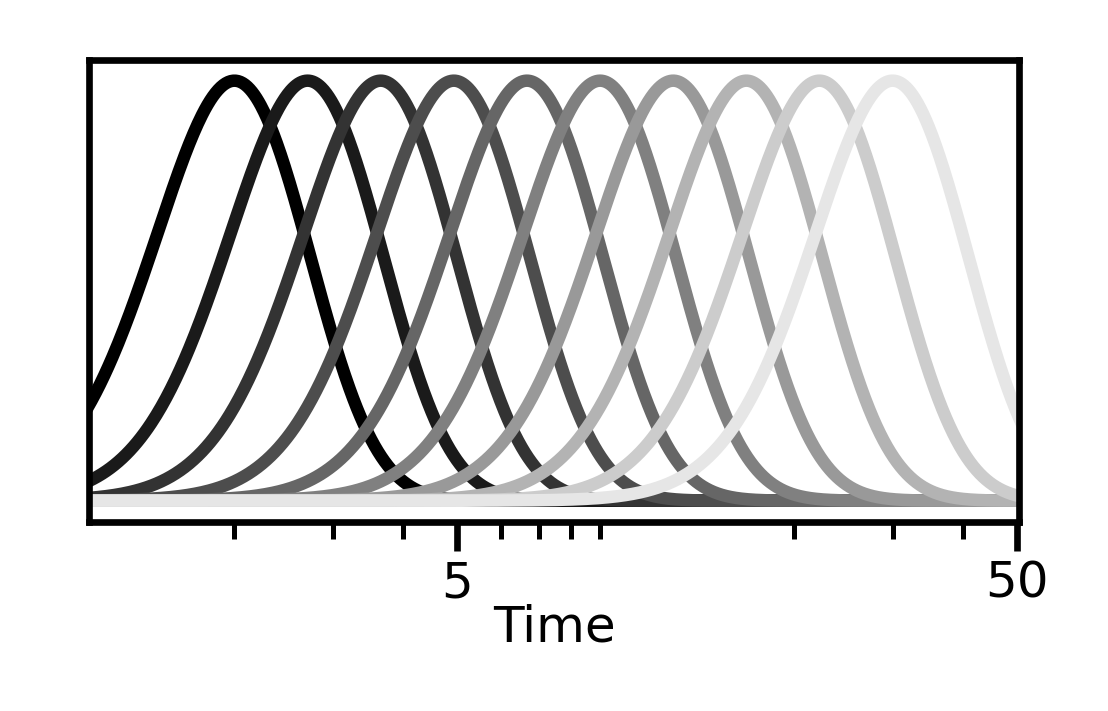}} &
        {\includegraphics[width=0.31\textwidth]{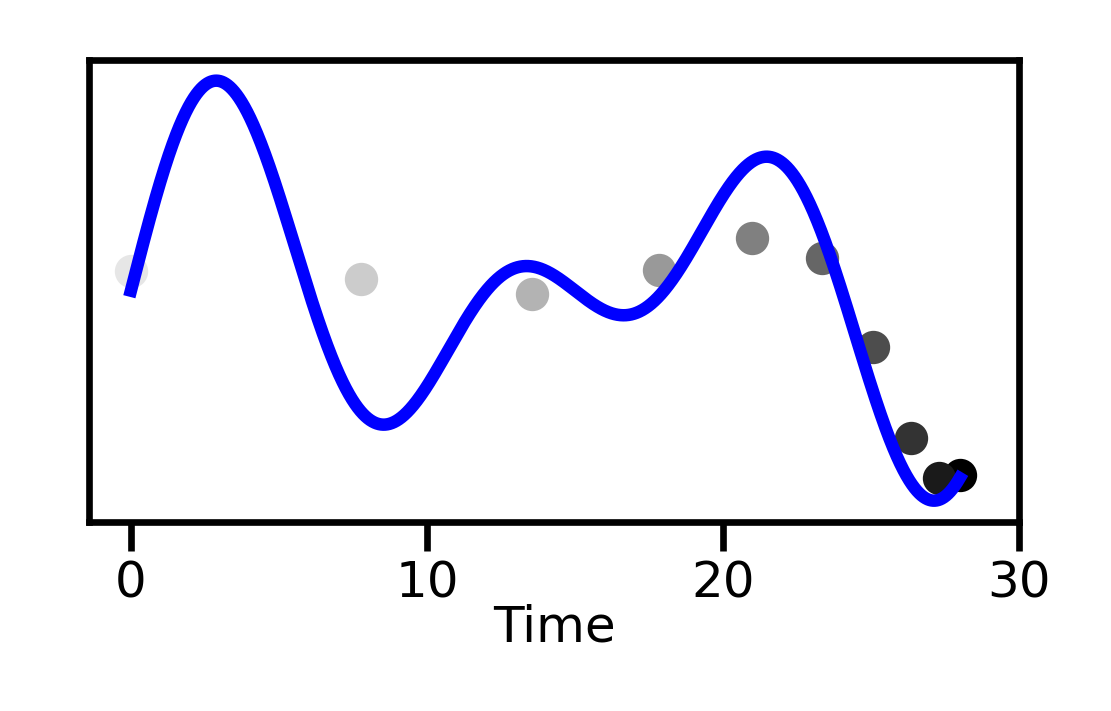}} 
    \end{tabular}
    \begin{tabular}{lll}
    \textbf{G} &
    \textbf{H} &
    \textbf{I} \\
        {\includegraphics[width=0.31\textwidth]{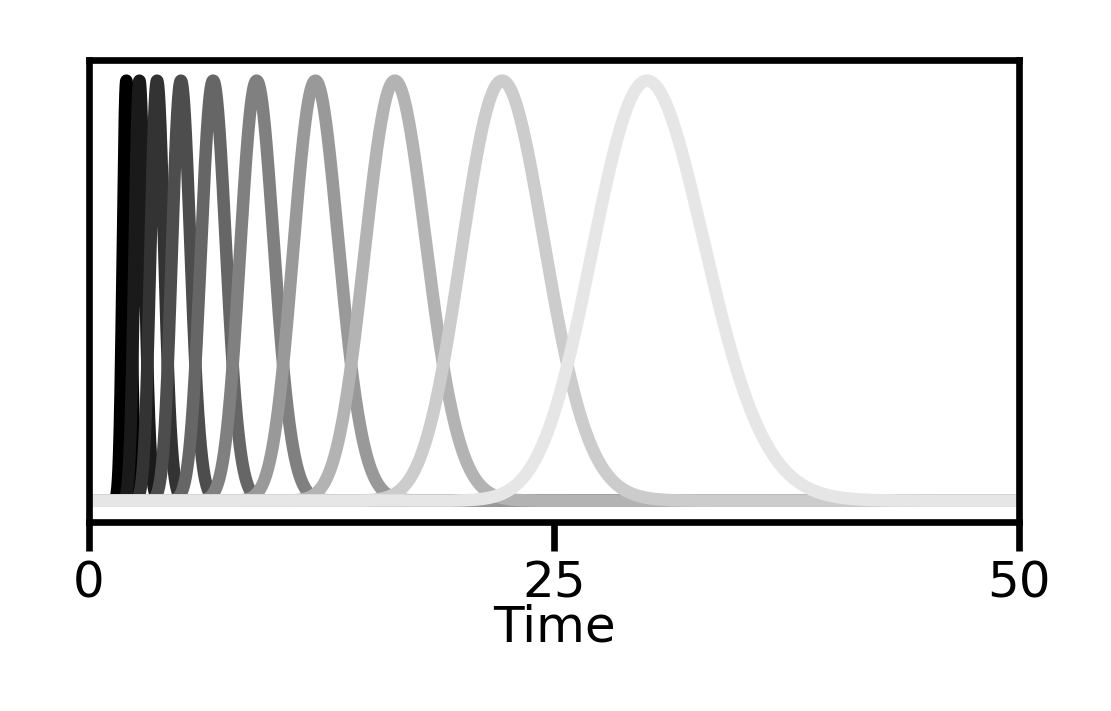}} &
        {\includegraphics[width=0.31\textwidth]{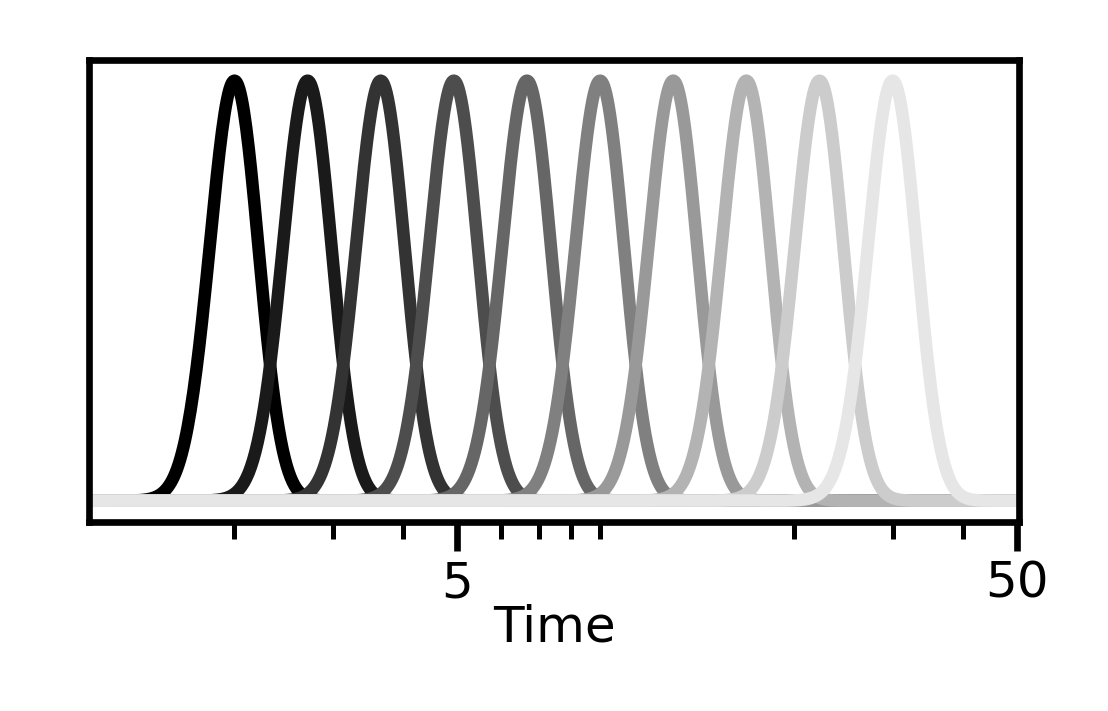}} &
        {\includegraphics[width=0.31\textwidth]{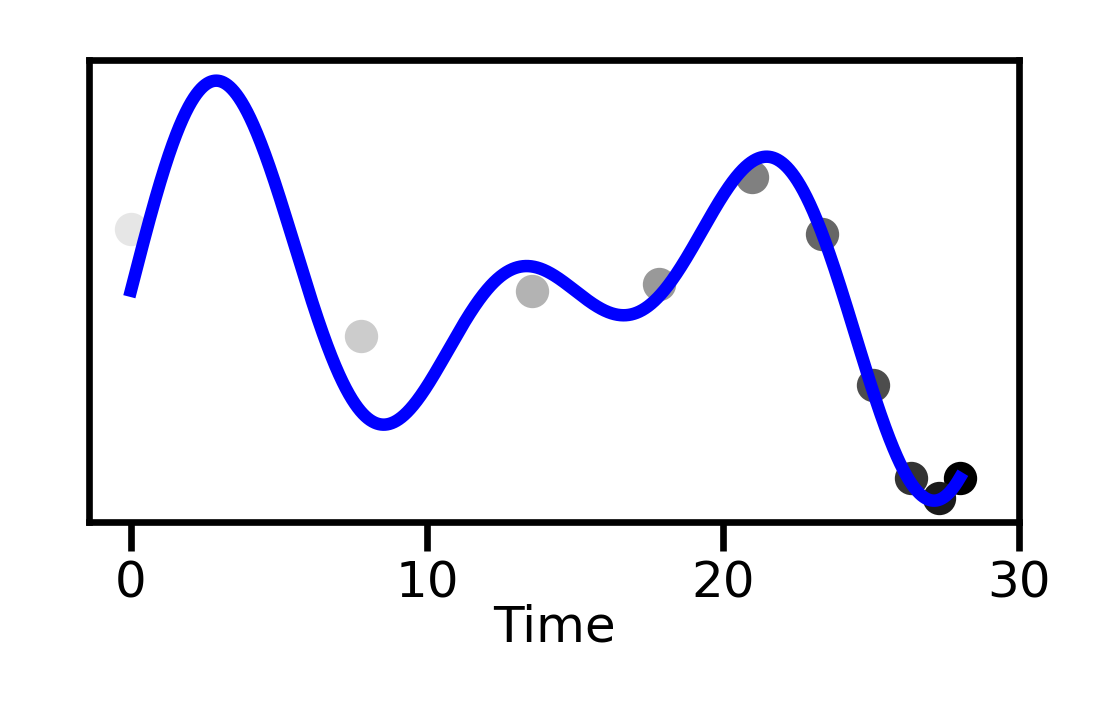}} 
    \end{tabular}
    \caption{Continuation of Fig.~\ref{fig:imp_res_main}. \textbf{D, E, F.} Same as panels A, B, C in the main text, but for $k=10$. Note that a lower value of $k$ results in wider filters but does not break the scale-invariance. Wider filters smooth the signal more, making the memory representation less able to follow changes in the signal (panel \emph{F}). \textbf{G, H, I.} Same as panels A, B, C, but for $k=100$. For very large values of $k$, the filters become narrower, making the space between them larger. Tokens that fall between peaks of the filters get less well represented than tokens that fall closer to the peaks.}
    \label{fig:imp_res_appendix}
\end{figure}

\begin{table}[!h]
\small
\centering
\begin{tabular}{c@{\hspace{8pt}}c@{\hspace{8pt}}c@{\hspace{8pt}}c@{\hspace{8pt}}c@{\hspace{8pt}}c}
\toprule
\textbf{\# Filters} & \textbf{Delta Filters} & \textbf{Filter Window} & \multicolumn{3}{c}{\textbf{Scale-Invariant Compression Filters}} \\
\cmidrule(lr){2-2} \cmidrule(lr){4-6}
& \emph{PPL} & ($\taustar_{max}$) & \textbf{$k=100$} & \textbf{$k=150$} & \textbf{$k=200$} \\
& \textbf{Raw / Per-Word}& & \textbf{Raw / Per-word} & \textbf{Raw / Per-word} & \textbf{Raw / Per-word} \\
\midrule
5  & 19.26/30.78 & 2    & 19.74/31.31 & 19.71/31.26 & 19.77/31.37 \\
9  & 19.14/30.44 & 4    & 19.46/30.80 & 19.45/30.80 & 19.47/30.82 \\
13 & 19.02/30.10 & 8    & 19.03/28.46 & 19.00/28.41 & 19.20/28.75 \\
17 & 18.89/29.76 & 16   & 18.97/30.87 & 18.75/30.45 & 18.88/30.69 \\
21 & 18.67/28.92 & 32   & 18.56/28.72 & 18.47/28.56 & 18.52/28.64 \\
25 & 18.70/29.79 & 64   & 18.23/28.91 & 18.24/28.93 & 18.34/29.13 \\
29 & 19.03/28.82 & 128  & 18.29/27.54 & 18.26/27.49 & 18.31/27.58 \\
33 & 18.29/28.67 & 256  & 17.56/27.35 & 17.58/27.39 & 17.48/27.21 \\
37 & 17.91/25.37 & 512  & 17.36/24.51 & 17.19/24.24 & 17.13/24.14 \\
41 & 18.12/29.89 & 1024 & 17.18/28.07 & 17.20/28.12 & 17.09/27.90 \\
45 & 17.97/27.41 & 2048 & 17.00/25.72 & 17.03/25.78 & 17.07/25.85 \\
49 & 17.96/28.08 & 4096 & 17.02/26.38 & 17.00/26.35 & 17.02/26.38 \\
53 & 17.98/25.48 & 8192 & 16.89/23.76 & 16.86/23.71 & \textbf{16.76/23.56} \\
\bottomrule
\end{tabular}
\vspace{4pt}
\caption{WikiText-103 perplexity on the test set for an uncompressed size of 256. \emph{Baseline} represents delta pulse filters where no compression occurs (control model). Our model, based on scale-invariant compression filters, was evaluated with $k=100$, $k=150$, and $k=200$. Note that $\taustar_{max}$ value is relevant only for our scale-invariant compression model. Each column reports two metrics in the format PPL / Per-Word PPL: PPL is the raw perplexity computed using the GPT-2 tokenizer, and Per-Word PPL is computed as \( e^{\text{Loss}_{tot} / n_{\text{words}}} \) where \emph{\( \text{Loss}_{\text{tot}} \)} is the total cross-entropy loss over the tokenized test set and $n_\text{words}$ is the number of words in the test set.}

\label{tab:wiki103}
\end{table}

\begin{table}[!h]
\small
\centering
\begin{tabular}{c@{\hspace{8pt}}c@{\hspace{8pt}}c@{\hspace{8pt}}c@{\hspace{8pt}}c@{\hspace{8pt}}c}
\toprule
\textbf{\# Filters} & \textbf{Delta Filters} & \textbf{Filter Window} & \multicolumn{3}{c}{\textbf{Scale-Invariant Compression Filters}} \\
\cmidrule(lr){2-2} \cmidrule(lr){4-6}
& \emph{PPL} & ($\taustar_{max}$) & \textbf{$k=100$} & \textbf{$k=150$} & \textbf{$k=200$} \\
& \textbf{Raw / Per-Word}& & \textbf{Raw / Per-word} & \textbf{Raw / Per-word} & \textbf{Raw / Per-word} \\
\midrule
5  & 21.00/102.11 & 2    & 21.05/101.66 & 21.08/101.90 & 21.11/102.14 \\
9  & 20.86/101.04 & 4    & 20.86/100.29 & 20.92/100.72 & 20.81/99.91 \\
13 & 20.73/99.97 & 8    & 20.69/99.62 & 20.67/99.42 & 20.69/99.61 \\
17 & 20.59/98.90 & 16   & 20.35/96.77 & 20.60/98.60 & 20.36/96.83 \\
21 & 20.45/97.83 & 32   & 20.46/96.87 & 20.40/97.43 & 20.65/99.28 \\
25 & 20.96/101.75 & 64   & 20.68/99.65 & 20.67/99.56 & 20.74/100.11 \\
29 & 22.16/110.45 & 128  & 21.36/104.48 & 21.40/104.74 & 21.44/105.09 \\
33 & 19.80/92.66 & 256  & 19.10/87.72 & 19.02/87.21 & 19.37/89.67 \\
37 & 19.39/90.12 & 512  & 18.69/85.26 & 18.75/85.64 & 18.82/86.14 \\
41 & 19.44/90.11 & 1024 & 18.68/84.81 & 18.57/84.02 & 18.46/83.23 \\
45 & 19.05/87.83 & 2048 & 17.97/80.41 & 17.86/79.64 & \textbf{17.82/79.40} \\
49 & 19.26/88.86 & 4096 & 18.07/80.66 & 18.15/81.18 & 18.35/82.56 \\
53 & 19.36/89.29 & 8192 & 18.39/82.60 & 18.25/81.62 & 18.18/81.16 \\
\bottomrule
\end{tabular}
\vspace{4pt}
\caption{PG-19 perplexity on test set after 4 epochs. The models have the same configuration as those used on WikiText-103.}
\label{tab:pg19}
\end{table}

\end{document}